\documentclass[runningheads]{llncs}

\usepackage[T1]{fontenc}
\usepackage{graphicx,verbatim}
\usepackage{tcolorbox}
\tcbuselibrary{raster, skins, breakable}
\usepackage{float}
\usepackage{multirow}
\usepackage{booktabs}
\usepackage{amsmath}
\tcbuselibrary{breakable}

\newcommand{\blfootnote}[1]{%
  \begingroup
  \renewcommand{\thefootnote}{}%
  \footnotetext{#1}%
  \endgroup
}

\begin{document}

\title{Spatial Message Passing in Language Space for Pathology Image Interpretation}
\titlerunning{Pathological Spatial Message Passing in Language Space}


\author{Jing-Cheng Yang\inst{1} \and
Hao-Jung Wang\inst{1} \and
Jinhao Du\inst{2} \and
Yang Hu\inst{2} \and
Ming-shan Tsai\inst{3} \and
Jens Rittscher\inst{2}* \and
Bin Li\inst{2}*}

\authorrunning{J.-C. Yang et al.}

\institute{
National Taiwan University, Taipei, Taiwan, 
\email{jingcheng.yang2004@gmail.com}
\and
University of Oxford, Oxford, United Kingdom, 
\email{\{firstname.lastname\}@eng.ox.ac.uk}
\and
ZYTCA Limited, Oxford, United Kingdom
\\
* Corresponding authors
}

\maketitle
\blfootnote{\textit{Accepted at MICCAI 2026 Workshop (Oral).}}

\begin{abstract}
Multimodal Large Language Models (MLLMs) can generate pathological descriptions from histological images, but gigapixel Whole Slide Images (WSIs) exceed their visual context limits. 
The standard tiling workaround makes WSIs tractable yet severs the tissue neighborhoods that define tumor–stroma interfaces and morphology.
We introduce Spatial Language Message Passing (SLMP), a framework that performs spatial reasoning entirely in language space, human-readable by construction. SLMP represents a WSI region as a spatial text graph: tiles are nodes initialized with MLLM descriptions, and edges encode spatial adjacency. 
For each tile, an LLM refines its description by integrating language messages from adjacent tiles under a shared aggregation policy that, on the tile grid, acts as an adaptive local kernel operating on text rather than learned embeddings. 
This policy is an inspectable prompt that can be refined from model-observed tissue phenotypes via textual gradients, enabling automatic semantic optimization from local cellular context to broader tissue morphology without fine-tuning MLLM weights. 
On representative HER2 and CAMELYON16 regions, SLMP improves tile-level tumor description accuracy in settings spanning general-purpose and pathology-specialized backbones, with gains of +3.3 to +19.6 percentage points. 
Random-neighbor ablations confirm that these gains stem from spatial context rather than additional text alone, and inspecting the optimized policies reveals interpretable, tissue-specific decision rules. 
Besides, without any weight updates or fine-tuning the backbone MLLM, SLMP substantially improves general-purpose MLLMs and narrows its gap to pathology-specialized counterparts, offering a transparent and flexible mechanism for incorporating spatial reasoning into MLLM-based pathology analysis.

\end{abstract}

\section{Introduction}
\label{sec:intro}

Digital pathology has transformed glass slides into Whole Slide Images (WSIs), enabling computational pathology systems to analyze tissue morphology at scale~\cite{Bera19DigitalPathology}. 
WSIs are gigapixel fields in which diagnostic interpretation depends not only on local cellular appearance, but also on the spatial organization of neighboring tissue. 
Multimodal Large Language Models (MLLMs) can describe individual pathology tiles in natural language~\cite{Sun24PathAsst,Liang25WSILLAVA,Zhang25PathoR1}, but their limited visual context makes direct WSI-level reasoning difficult. 
A common workaround is to tile the slide, yet isolated tiles often remove the microenvironmental continuity needed to interpret tumor-stroma interfaces, lymphoid architecture, or boundary regions.


Restoring this lost spatial context is therefore the central challenge. 
Prevailing WSI methods reintroduce spatial structure, if at all, inside numerical feature spaces: Multiple Instance Learning and pathology foundation models embed each tile independently and pool the instances into a slide-level vector~\cite{Ilse18ABMIL,Xu24ProvGigaPath}, often without explicitly modeling local physical adjacency, while spatially aware variants model inter-tile relations using attentions over learned embeddings~\cite{Shao21TransMIL,Chen21PatchGCN}. These representations can be effective but typically require task-specific training and the interpretability remains limited. 
Agentic MLLM pipelines restore language-level reasoning~\cite{Hong26AdaptiveReasoning}, yet still interpret each tile in isolation. What is missing is a mechanism by which tile-level \emph{language} descriptions exchange local spatial evidence directly.

We propose \textbf{Spatial Language Message Passing (SLMP)}, a prompt-optimized framework for pathology image interpretation. SLMP represents a WSI region as a spatial text graph: each tile is a node initialized with an MLLM-generated morphology description, and edges connect physically adjacent tiles. For each center tile, an LLM updates the description by integrating messages from its cardinal neighbors under a single shared aggregation policy. 
On the regular WSI grid, this shared policy behaves like a convolutional kernel sliding across the lattice, but the representation and the aggregation rule are all expressed in natural language rather than numerical feature vectors. The prompt and the LLM therefore function as a language-space aggregation operator that determines how neighboring descriptions are weighted and integrated into a context-aware representation.
This neighborhood-aggregation principle echoes graph message
passing~\cite{Kipf17GCN,Velickovic18GAT} and recent attempts to prompt LLMs over text-attributed graphs~\cite{Wang23NLGraph,Zhu25LLMasGNN}, but those target sparse symbolic graphs, whereas SLMP operates on the dense, physically grounded tissue lattice and aggregates through a generative operator.  
Moreover, we optimize this prompt with TextGrad~\cite{Yuksekgonul24TextGrad}, 
which uses LLM-generated textual feedback as a gradient-like signal for improving compound language systems. In SLMP, these textual gradients identify failures in local evidence passing and revise the prompt-level aggregation policy while leaving all MLLM weights frozen. Because the policy is a language artifact rather than latent weights, it can be read directly, exposing the tissue-specific decision rules the system acquires.

Our contributions are threefold. 1) We formulate visual pathological interpretation as spatial message passing over tile-level language space, preserving neighborhood structure and spatial context. 
2) We instantiate the aggregation rule as a natural-language policy optimized via textual gradients without tuning LLM model weights, yielding an interpretable and readable message-passing policy that acts as a spatial adapter for pathological spatial reasoning.
3) On regions from HER2~\cite{Farahmand22HER2} and CAMELYON16 (C16)~\cite{Ehteshami17CAMELYON16} datasets, SLMP improves tile-level tumor classification across general-purpose and pathology-specialized backbones by up to $+19.6$ percentage points, and lets general-purpose GPT-4.1 family narrow the gap to specialized models without any weight updates.

\section{Methods}
\label{sec:methods}


\begin{figure}
    \centering
    \includegraphics[width=\textwidth]{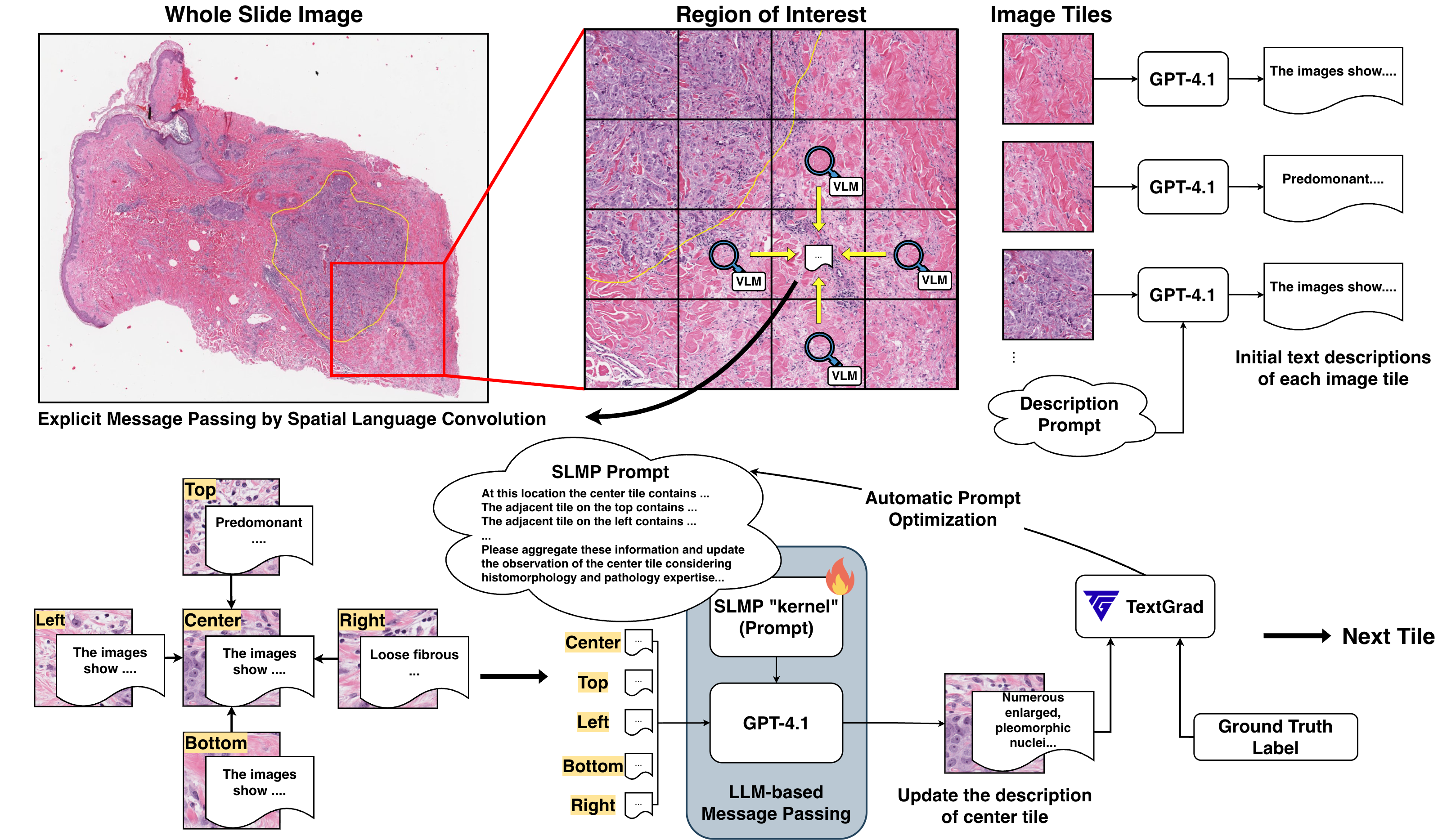}
    \caption{Overview of the proposed framework.
    \textbf{(Top)} A WSI region of interest is tiled and each tile
    is described independently by an MLLM.
    \textbf{(Bottom-left)} Tiles and their descriptions form a
    spatial text graph; the LLM kernel aggregates the cardinal
    neighbors to update the center-tile description.
    \textbf{(Bottom-right)} TextGrad propagates textual gradients
    back through the pipeline to optimize the kernel prompt
    $\mathbf{p}$.}
    \label{fig:overview}
\end{figure}

SLMP is a three-stage framework for adding local spatial context to MLLM-generated pathology tile descriptions. The method first converts image tiles into language, then treats these descriptions as node states on a spatial text graph, and finally learns a prompt-level aggregation policy that updates each node from its neighbors. Throughout the pipeline, all MLLM weights remain frozen; only the language prompt controlling message passing is optimized. 
A defining property of this design is that every intermediate state is natural language: the per-tile descriptions, the messages exchanged between tiles,
and the aggregation policy itself are all human-readable text. Spatial reasoning in SLMP is therefore inspectable by construction, rather than recovered post-hoc from saliency maps or attention weights over opaque embeddings.                                

As illustrated in Fig.~\ref{fig:overview}, our pipeline consists of three interconnected stages:
(1)~\textbf{Encoding}, where each tile is described independently by an MLLM;
(2)~\textbf{Spatial language message passing}, where an LLM updates each center-tile description using messages from adjacent tiles; and
(3)~\textbf{Prompt optimization}, where TextGrad refines the adaptive local aggregation rule.

\subsection{WSI Tiling and Initial Description Generation}

Given a Region of Interest (ROI) from a WSI, we partition the tissue into a regular $N{\times}N$ grid of non-overlapping tiles.  The regular grid is important: it preserves the physical adjacency structure that will later define message passing edges. 
 We evaluate on WSI regions rather than whole slides to bound the cost of per-tile MLLM inference. The selected ROIs nonetheless span tumor, stroma, and tumor-adjacent normal tissue, exercising exactly the cross-tile interfaces that SLMP targets.

Each tile $v_i$ is processed independently by an MLLM to produce an initial description $d_i^{(0)}$. The prompt can ask for morphology, tissue architecture, staining, and diagnostically relevant visual cues. No neighboring information is provided at this stage, so $d_i^{(0)}$ reflects the tile's local evidence alone. We generate these descriptions using both general-purpose and pathology-specialized backbones, 
allowing SLMP to be evaluated as a spatial reasoning layer over different descriptor distributions.

\begin{figure}[!t]
    \centering
    \includegraphics[width=\textwidth]{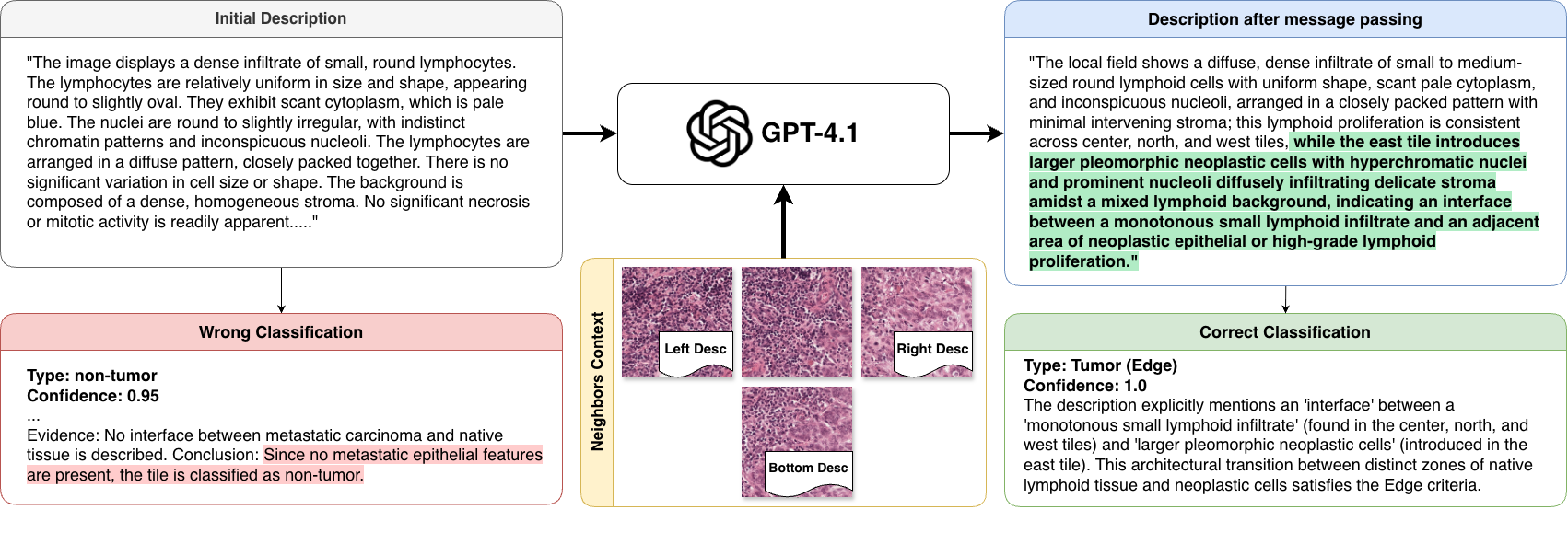}
    \caption{A representative tile from CAMELYON16 where spatial context corrects an initially wrong classification. The center tile's standalone description (left) lacks evidence of malignancy, yielding a confident non-tumor prediction. After message passing, the updated description integrates evidence from the east neighbor---where larger pleomorphic neoplastic cells are visible---correctly identifying a tumor-normal interface.}
    \label{fig:message_passing}
\end{figure}

\subsection{Spatial Language Message Passing}

We represent the tile grid as a spatial text graph $\mathcal{G}=(\mathcal{V},\mathcal{E})$. Each node $v_i\!\in\!\mathcal{V}$ stores a natural-language description $d_i$, and edges $\mathcal{E}$ encode 4-connectivity: top, bottom, left, and right neighbors. For a center tile $v_c$, SLMP applies an LLM governed by prompt $\mathbf{p}$ to the center description and its neighbor descriptions. The LLM returns an updated description $d_c^{(t+1)}$ that preserves local evidence while incorporating spatially adjacent morphology (Fig.~\ref{fig:message_passing}):
\begin{equation}
  d_c^{(t+1)} \;=\;
  \mathrm{LLM}_{\mathbf{p}}\!\!\left(
      d_c^{(t)},\;
      d_{\mathrm{top}}^{(t)},\;
      d_{\mathrm{bot}}^{(t)},\;
      d_{\mathrm{left}}^{(t)},\;
      d_{\mathrm{right}}^{(t)}
  \right).
  \label{eq:conv}
\end{equation}

For border tiles the neighbor set contains only the 2--3 available neighbors; the same prompt handles these cases, and boundary handling is itself one of the rules exposed to optimization.
This update is applied at every tile in the grid with the same prompt-level aggregation policy. Because the graph is a regular image lattice, the operation is analogous to a convolutional kernel sliding across a pixel map; however, the representation, messages, and aggregation rule are expressed in language rather than numerical feature vectors. 
The key difference is that conventional convolutional kernels carry fixed pretrained numerical weights, whereas our language-space kernel is adaptive: the prompt $\mathbf{p}$ and the LLM jointly define a generative local operator that interprets the surrounding tissue context. Rather than applying static weights to neighboring features, this kernel reads the center tile and its neighbors, weighs repeated evidence, suppresses isolated outliers, reconciles conflicting neighbor signals, and rewrites the center description to capture cross-tile morphology such as stromal invasion fronts, glandular continuity, and nuclear pleomorphism.
Fig.~\ref{fig:message_passing} shows a concrete trace of this operation: the standalone description, the neighbor evidence that is integrated, and the corrected output are all legible to a pathologist, making the spatial inference verifiable rather than merely accurate.

\subsection{Prompt Optimization via Textual Gradients}

Since all model weights are frozen, SLMP optimizes only the prompt $\mathbf{p}$ that controls the local aggregation policy. We write the one-step update from Eq.~\ref{eq:conv} as an explicitly prompt-conditioned operator:
\begin{equation}
d_i^{(1)} =
\mathrm{SLMP}\left(
d_i^{(0)}, \{d_j^{(0)} : j \in \mathcal{N}_i\}; \mathbf{p}
\right)
\label{eq:prompt-conditioned-slmp}
\end{equation}
where $\mathcal{N}_i$ denotes the spatial neighbors of tile $i$. The goal is to find a prompt that produces descriptions whose neighbor integration improves downstream tile interpretation:
\begin{equation}
\mathbf{p}^{*} =
\arg\min_{\mathbf{p}} \mathcal{L}(\mathbf{p}), \quad
\mathcal{L}(\mathbf{p}) =
\sum_{i \in \mathcal{T}}
\mathrm{Eval}\!\left(d_i^{(1)}(\mathbf{p}), \mathcal{N}_i, y_i\right).
\label{eq:prompt-optimization}
\end{equation}

Here, $\mathrm{Eval}$ checks (i) whether $d_i^{(1)}$ is consistent with the ground-truth label $y_i$ and (ii) whether neighbor context was reflected in the update. Eq.~\ref{eq:prompt-optimization} states the objective conceptually; optimization is driven by the evaluator's structured textual feedback rather than a scalar gradient.
Cases where no neighbor evidence appears are labeled \textit{no passing}; otherwise the forward pass is categorized as \textit{helpful context}, \textit{confirmation}, \textit{hallucination}, \textit{wrong neighbor focus}, \textit{over-interpretation}, or \textit{under-interpretation}.

Textual gradients are produced by a fault-attribution policy that first determines whether an error stems from a difficult input or from the prompt itself. Prompt errors are further classified as a \textit{policy mistake} (an encoded rule is wrong or unsuitable), a \textit{phrasing mistake} (ambiguous wording causes misinterpretation), or an \textit{edge-case gap} (the prompt lacks coverage for a rare input type). Each diagnosis triggers a targeted revision: bolder rules, rewording, or integration of specific pathology terminology, respectively.
Because both the evaluation outcome and the fault attribution are expressed as discrete, named categories, every optimization step carries an auditable rationale: one can read why a given prompt revision was made and trace how the policy evolves across iterations. The optimization trajectory is thus itself an object of inspection rather than a black-box parameter update.

\section{Experiments and Results}
\label{sec:results}

\subsection{Experimental Setup}

We evaluate SLMP on HER2 breast cancer ROIs~\cite{Farahmand22HER2} and C16 lymph node ROIs~\cite{Ehteshami17CAMELYON16}. Each WSI region is tiled into a $7{\times}7$ grid at $214 \times 214\,\mu\text{m}$ per tile. ROIs are selected to contain both tumor and normal tissue, excluding regions with tumor in only a single tile. This yields 40 HER2 and 25 C16 regions with tile-level annotations; five HER2 and seven C16 regions are used for prompt optimization and the remainder held out for validation.

Four backbones generate initial descriptions $d_i^{(0)}$: GPT-4.1, PathGEN-LLaVA 27B, Patho-R1 7B, and MedGemma-1.5 4B. SLMP performs a single round of message passing ($T{=}1$) with $\mathbf{p}^{*}$. Neighbor context is supplied as a JSON object with cardinal-direction keys; boundary tiles set missing neighbors to \texttt{None}. Diagnostic utility is measured by a frozen GPT-4.1-mini judge predicting tile-level tumor presence from the description alone---a proxy for whether the description is informationally complete and spatially precise. Generation uses temperature $1$; the judge uses temperature $0$. We report mean $\pm$ std over three independent evaluation runs to capture small judge fluctuations.

\textbf{Initial Aggregation Prompt.}
Optimization begins from a single $\mathbf{p}_{\text{initial}}$ shared across both datasets before any specialization (Sec.~\ref{sec:ablation-specialization}). It encodes generic spatial reasoning rules with no pathology-specific vocabulary; tissue-specific criteria emerge entirely through subsequent optimization.
\begin{tcolorbox}[
  colback=gray!5, colframe=gray!45,
  title={$\mathbf{p}_{\textbf{initial}}$ (excerpt)}]
\scriptsize\ttfamily
\ldots \\
\textbf{SEMANTIC REASONING POLICY}\\
1. \textbf{THE HOMOGENEITY RULE:} If the center and the majority of neighbors share the same tissue type (e.g., all dense tumor), reinforce the center's confidence and describe it as a ``continuous solid region.'' \\[2pt]
2. \textbf{THE OUTLIER RULE:} If the center contains a minor, ambiguous feature but all neighbors are definitively empty or benign, treat the center as likely noise/benign context\ldots \\[2pt]
\ldots
\end{tcolorbox}

\begin{table}[t]
\caption{Tile-level tumor classification accuracy before and after SLMP. 
PathGEN is a 27B pathology-specialized model; Patho-R1 is 7B; MedGemma-1.5 is 4B; GPT-4.1 is a general-purpose model without pathology pretraining.}\label{tab:avg}
\centering
\small
\resizebox{0.9\textwidth}{!}{%
\begin{tabular}{llccc}
\hline
Dataset & Base Model & Before & After & $\Delta$ \\
\hline
\multirow{4}{*}{HER2}
  & GPT-4.1         & 0.6163 $\pm$ 0.0049 & 0.7338 $\pm$ 0.0027 & \textbf{+11.75 pp} \\
  & PathGEN         & 0.6959 $\pm$ 0.0049 & 0.7674 $\pm$ 0.0033 & \textbf{+7.15 pp}  \\
  & Patho-R1        & 0.4695 $\pm$ 0.0005 & 0.5267 $\pm$ 0.0005 & \textbf{+5.72 pp}  \\
  & MedGemma-1.5    & 0.6738 $\pm$ 0.0025 & 0.7071 $\pm$ 0.0010 & \textbf{+3.33 pp}         \\
\hline
\multirow{4}{*}{CAMELYON16}
  & GPT-4.1         & 0.7422 $\pm$ 0.0040 & 0.7994 $\pm$ 0.0031 & \textbf{+5.72 pp}  \\
  & PathGEN         & 0.6511 $\pm$ 0.0040 & 0.8186 $\pm$ 0.0040 & \textbf{+16.75 pp} \\
  & Patho-R1        & 0.5397 $\pm$ 0.0015 & 0.7361 $\pm$ 0.0038 & \textbf{+19.64 pp} \\
  & MedGemma-1.5    & 0.5045 $\pm$ 0.0026 & 0.6728 $\pm$ 0.0053 & \textbf{+16.8 pp}  \\
\hline
\end{tabular}
}
\end{table}

\subsection{Main Results}
Table~\ref{tab:avg} reports tile-level tumor classification accuracy before and after SLMP. All eight model--dataset settings improve, by +3.3 to +19.6 pp, with the largest gains on C16. Neighbor-aware language updates thus improve diagnostic utility across both general-purpose and pathology-specialized descriptors.

\begin{figure}[t]
    \centering
    \includegraphics[width=0.99\textwidth]{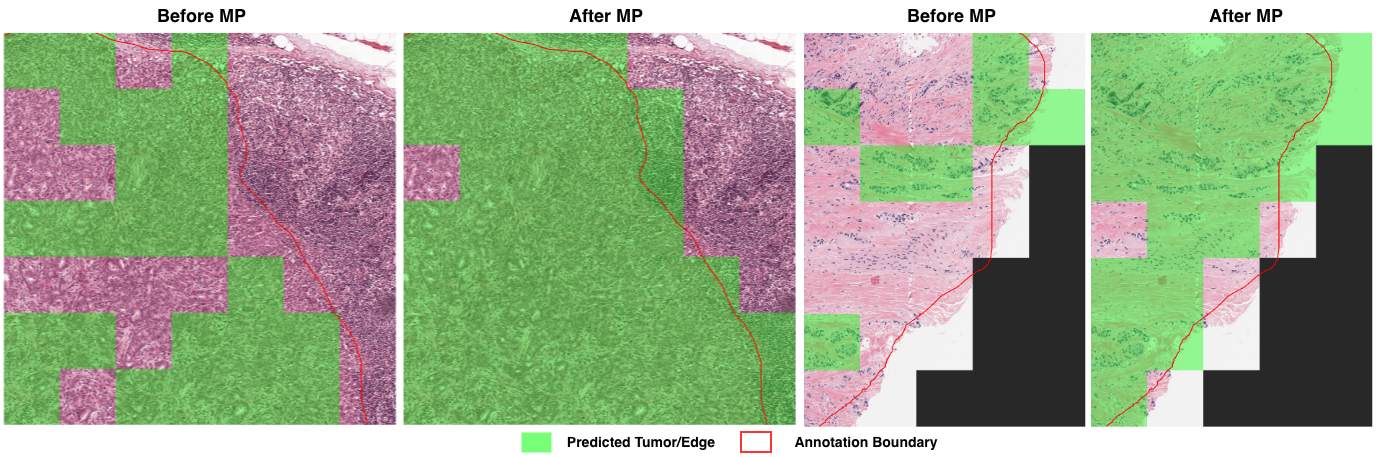}
    \caption{Tile-level prediction maps on $7{\times}7$ ROIs. \textbf{(Left)}~CAMELYON16. \textbf{(Right)}~HER2. SLMP consolidates fragmented predictions into coherent fields.}
    \label{fig:heatmaps}
\end{figure}

Qualitatively, SLMP changes the spatial pattern of predictions rather than merely correcting isolated labels, as shown in Fig.~\ref{fig:heatmaps}. Before message passing(MP), tile predictions are fragmented, with false negatives appearing inside tumor regions and uncertain boundary tiles. After SLMP, predictions consolidate into more coherent fields: tumor tiles cluster, and tumor-normal boundaries become more structurally defined. This suggests that the updated descriptions recover local spatial context that is lost when tiles are interpreted independently.

\subsection{Ablation Study}


To test whether gains come from spatial context rather than extra text, we replace each tile's true neighbors with randomly sampled tiles from the same ROI, preserving text volume but removing physical adjacency. Table~\ref{tab:ablation} compares \emph{No Message Passing (Baseline)}, \emph{SLMP}, and \emph{Random Neighbors} on a randomly sampled subset of 8 regions (initial tile descriptions generated using GPT-4.1) per dataset from HER2 and C16.

True neighbors consistently outperform random neighbors: on C16, random neighbors drop accuracy \emph{below} the No Message Passing baseline (0.6250 vs.\ 0.7358), while SLMP reaches 0.7830; on HER2 the same pattern holds.

We further isolate prompt optimization by comparing $\mathbf{p}_{\text{initial}}$ against $\mathbf{p}^{*}$ with GPT-4.1 on a separate randomly sampled subset of 8 regions per dataset(Table~\ref{tab:ablation-prompt}). On C16, $\mathbf{p}_{\text{initial}}$ \emph{underperforms} No Message Passing baseline (0.6981 vs.\ 0.7547), showing that naive aggregation without a tissue-aware policy may actively harm predictions; $\mathbf{p}^{*}$ recovers to 0.8121. Together, these ablations confirm that SLMP requires both spatially coherent neighbors and an optimized policy.

\begin{table}
\centering
\caption{Spatial neighbor ablation. SLMP uses true spatial neighbors; Random substitutes randomly sampled tiles from the same ROI. Mean $\pm$ std over three runs.}
\label{tab:ablation}
\small
\begin{tabular}{llccc}
\hline
Dataset & Metric & Baseline & Random & SLMP \\
\hline
\multirow{2}{*}{HER2}
  & Accuracy & 0.7153 & $0.6762 \pm 0.0098$ & $0.7727 \pm 0.0059$ \\
  & F1       & 0.6361 & $0.6982 \pm 0.0126$ & $0.7678 \pm 0.0016$ \\
\hline
\multirow{2}{*}{CAMELYON16}
  & Accuracy & 0.7358 & $0.6250 \pm 0.0069$ & $0.7830 \pm 0.0153$ \\
  & F1       & 0.7157 & $0.6831 \pm 0.0032$ & $0.8001 \pm 0.0124$ \\
\hline
\end{tabular}
\end{table}

\begin{table}
\centering
\caption{Prompt ablation. We compare message-passing results using the initial prompt $\mathbf{p}_{\text{initial}}$ (before optimization) against the optimized prompt $\mathbf{p}^{*}$ (after convergence). Mean $\pm$ std over three runs.}
\label{tab:ablation-prompt}
\small
\begin{tabular}{llccc}
\hline
Dataset & Metric & Baseline & Initial Prompt & Optimized Prompt \\
\hline
\multirow{2}{*}{HER2}
  & Accuracy & 0.7225 & $0.7416 \pm 0.0000$ & $0.7982 \pm 0.0049$ \\
  & F1       & 0.6463 & $0.7075 \pm 0.0037$ & $0.7787 \pm 0.0046$ \\
\hline
\multirow{2}{*}{CAMELYON16}
  & Accuracy & 0.7547 & $0.6981 \pm 0.0033$ & $0.8121 \pm 0.0011$ \\
  & F1       & 0.7615 & $0.7690 \pm 0.0021$ & $0.8257 \pm 0.0009$ \\
\hline
\end{tabular}
\end{table}




\subsection{Dataset-Specific Kernel Specialization}
\label{sec:ablation-specialization}


\begin{tcbraster}[
  raster columns=20,
  raster equal height=rows,
  raster column skip=0.02\textwidth,
  raster left skip=0pt,
  raster right skip=0pt,
]
\begin{tcolorbox}[
  colback=orange!5, colframe=orange!45,
  raster multicolumn=9,
  title={$\mathbf{p}^{*}_{\textbf{HER2}}$ (excerpt)}]
\scriptsize\ttfamily
\ldots \\
- Interpret spindle cell morphology in neighbors as possibly
  reflecting tumor heterogeneity or carcinoma variants with
  spindle features, \textbf{not exclusively sarcoma}. \\[2pt]
- Biphasic or heterogeneous patterns, including transitions
  in cellularity or matrix composition. \\[2pt]
- Evaluate all tiles for epithelial, glandular, spindle cell,
  stromal, inflammatory, and cytologic features, including
  nuclear pleomorphism, hyperchromasia, mitotic activity,
  architectural disruption, necrosis, and subtle stromal
  changes. \\[2pt]
\ldots
\end{tcolorbox}
\begin{tcolorbox}[
  colback=blue!5, colframe=blue!35,
  raster multicolumn=11,
  title={$\mathbf{p}^{*}_{\textbf{CAM}}$ (excerpt)}]
\scriptsize\ttfamily
\ldots \\
- Use a \textbf{diffuse lymphoid-appearing field} only when
  most tiles show: \textbf{uniform small dark round-to-oval
  cells}, \textbf{scant pale cytoplasm}, \textbf{inconspicuous
  nucleoli}, \textbf{preserved diffuse spacing}. \\[2pt]
- Do not let sparse atypia, collagen, adipose tissue, debris,
  \textbf{follicles}, or \textbf{germinal centers} dominate
  unless they are part of the repeated field-wide pattern. \\[2pt]
- If 2 or more tiles show tightly packed pink acini/tubules,
  cohesive nests/cords/sheets, crowded polygonal or
  round-to-oval epithelial cells, hyperchromatic or pleomorphic
  nuclei, visible nucleoli, scant cytoplasm, or loss of
  gland/lumen organization, describe a \textbf{solid or
  gland-forming neoplastic epithelial field}. \\[2pt]
\ldots
\end{tcolorbox}
\end{tcbraster}
Starting from the same $\mathbf{p}_{\text{initial}}$, HER2 and C16 optimization converge to distinct pathological vocabularies and decision rules, confirming that SLMP learns tissue-specific policies rather than generic rephrasing. The resulting rules can be read directly rather than inferred from saliency.

\subsection{Prompt Evolution Analysis}

In the C16-trained prompt, early iterations rely on generic homogeneity and outlier rules; by mid-optimization, pathological overrides emerge; at convergence, the policy grounds decisions in quantitative criteria such as repeated-evidence thresholds across neighboring tiles. Textual gradients thus do more than polish wording; they specialize a generic spatial prompt into a tissue-aware policy.

\begin{tcbraster}[
  raster columns=20,
  raster equal height=rows,
  raster column skip=0.02\textwidth,
  raster left skip=0pt,
  raster right skip=0pt,
]
\begin{tcolorbox}[
  raster multicolumn=9,
title={$\mathbf{p}_{\textbf{v30}}$ (excerpt)}]
\scriptsize\ttfamily
\ldots \\
- \textbf{Hard override for epithelial / carcinoma-like neighbors}\\
  If \textbf{any neighbor} shows pleomorphic enlarged cells, eosinophilic or pink cytoplasm, increased nuclear-to-cytoplasmic ratio\ldots then update toward a \textbf{tumor-edge / infiltrative epithelial interface}. Do \textbf{not} keep a bland lymphoid interpretation if these features are present\ldots \\
\end{tcolorbox}
\begin{tcolorbox}[
  raster multicolumn=11,title={$\mathbf{p}_{\textbf{v67}}$ (excerpt)}]
\scriptsize\ttfamily
\ldots \\
- \textbf{Merge repeated evidence; suppress isolated outliers}\\
  Treat features seen in only one tile as \textbf{contextual noise} unless they repeat in \textbf{2+ tiles}. \\[2pt]
- \textbf{Strong override for epithelial/glandular architecture}\\
  If \textbf{2 or more tiles} show tightly packed pink acini/tubules\ldots describe a \textbf{solid or gland-forming neoplastic epithelial field}. \textbf{Neighbor epithelial architecture overrides a lymphoid-looking center}. \\[2pt]
\end{tcolorbox}
\end{tcbraster}

\section{Discussion}
\label{sec:conclusion}
We presented SLMP, a prompt-optimized message-passing layer that adds local tissue context to MLLM descriptions of pathological images. By treating tile descriptions as node states on a spatial text graph, SLMP updates each center tile from its true neighbors while keeping all MLLM weights frozen, improving tile-level tumor classification in all eight model-dataset settings. Because the messages and the learned policy are natural language, this spatial reasoning stays human-readable and auditable.
SLMP thus offers a lightweight route to structured spatial reasoning in MLLMs: not by retraining the model, but by teaching language descriptions how to pass pathology evidence. Our study is nonetheless limited to curated ROIs rather than whole slides, and uses a single message-passing round: each round adds one
generative pass per tile over a describe-once baseline, which---unlike
embedding-based pooling---does not amortize.
A natural next step is visual-language message passing, where messages also carry the image tiles for direct visual integration.

\section*{Disclosure of Interests}

Ming-shan Tsai is a co-founder of ZYTCA Ltd. ZYTCA Ltd. provided
in-kind support for this work in the form of credits used to access OpenAI
services. The remaining authors declare that they have no competing interests
relevant to the content of this paper.

\bibliographystyle{splncs04}
\bibliography{Paper-0008}


\end{document}